\documentclass[journal]{IEEEtran}
\ifCLASSINFOpdf
  \usepackage[pdftex]{graphicx}
\else
\fi
\usepackage[T1]{fontenc} 
\usepackage{amsmath}
\usepackage[cmintegrals]{newtxmath}
\usepackage{bm} 
\usepackage{algorithmic}

\usepackage{array}

\usepackage{multirow}
\usepackage[table,xcdraw]{xcolor}
\usepackage{cite}

\begin{document}
%
\title{DFR: Deep Feature Reconstruction for Unsupervised Anomaly Segmentation }
%
%

\author{Jie Yang, Yong Shi, ZhiQuan Qi}
\maketitle

\begin{abstract}
Automatic detecting anomalous regions in images of objects or textures without priors of the anomalies is challenging, especially when the anomalies appear in very small areas of the images, making difficult-to-detect visual variations, such as defects on manufacturing products.
This paper proposes an effective unsupervised anomaly segmentation approach that can detect and segments out the anomalies in small and confined regions of images. Concretely, we develop a multi-scale regional feature generator which can generate multiple spatial context-aware representations from pre-trained deep convolutional networks for every subregion of an image.  
The regional representations not only describe the local characteristics of corresponding regions but also encode their multiple spatial context information, making them discriminative and very beneficial for anomaly detection.
Leveraging these descriptive regional features, we then design a deep yet efficient convolutional autoencoder and detect anomalous regions within images via fast feature reconstruction.
Our method is simple yet effective and efficient. It advances the state-of-the-art performances on several benchmark datasets and shows great potential for real applications.

\end{abstract}



\section{Introduction}
%
%
%
%


\IEEEPARstart{U}{supervised} anomaly segmentation aims at precisely detecting and localizing anomalous regions within images solely via prior knowledge from the anomaly-free images. This task is significant especially in smart manufacturing processes for ensuring qualified products, such as automatically inspecting and screening defective or flawed products. In these inspection scenarios, it is usually preferable to train machine learning models only with normal images of the products alone to detect the anomalies. Since industrial processes are generally optimized to produce least unqualified samples, it might be impossible to collect a sufficient amount or even a few of defective samples. More importantly, because all sorts of anomalies or defects would possibly occur during manufacturing, a detection model solely trained on limited anomaly samples may fail to generalize on those previously unseen ones.

\begin{figure}
	\centering
	\includegraphics[width=0.95\linewidth]{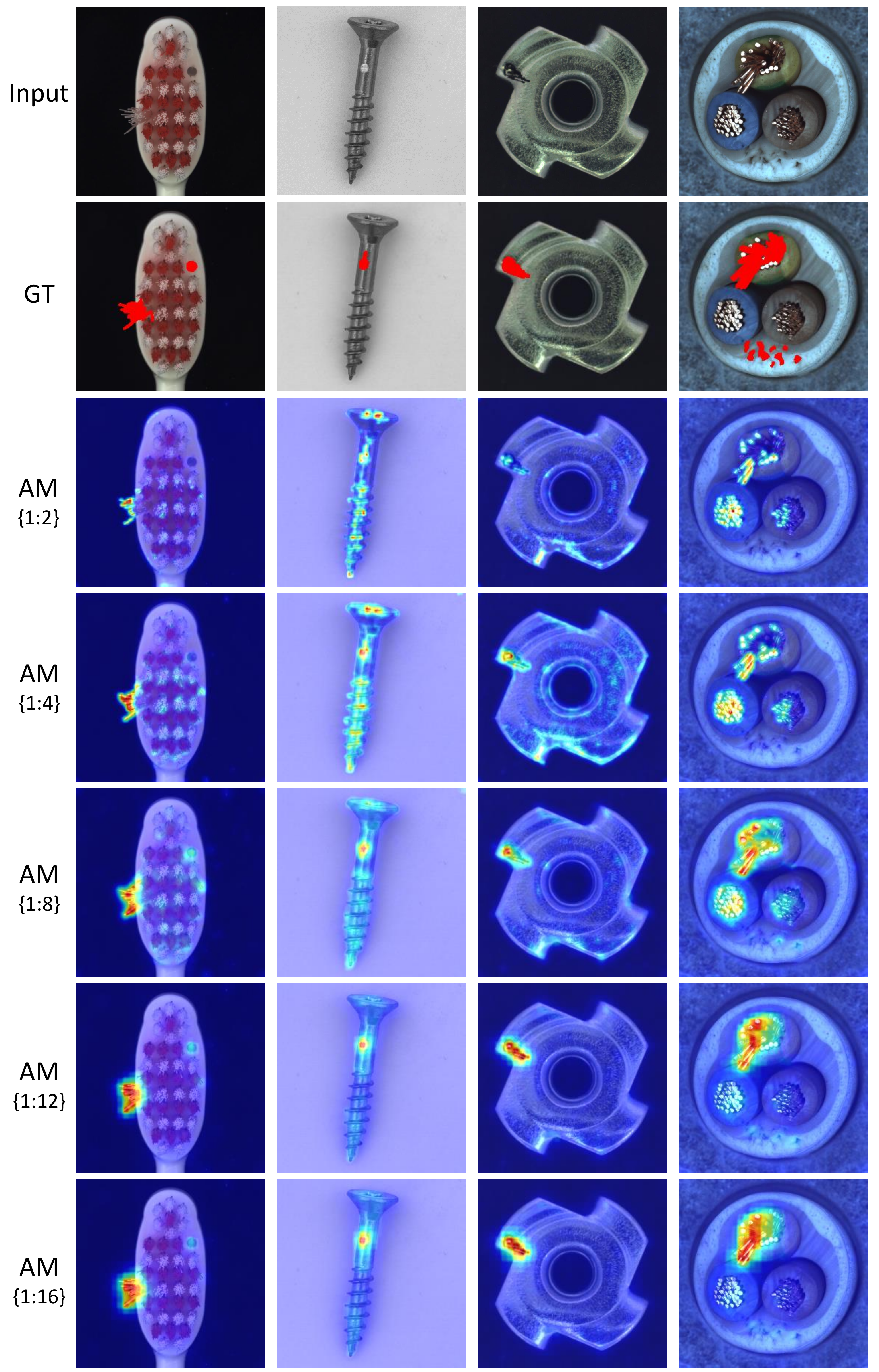}
	\caption{Qualitative results of our anomaly detection method with increasing feature scales on the MVTec AD dataset. Input: input image. AM \{1:$l$\}: anomaly map of our approach where representation scales from 1 to $l$ are leveraged. Note that in AM red regions correspond to high score for anomaly.}
	\label{multi-level}
\end{figure}
In recent literature, many effective unsupervised anomaly and novelty detection algorithms for images have been proposed~\cite{occ-tf, occ-tf-m, adv-one-class, ocgan, ae-mm, ae-lsa}, whereas most of these methods aim to image-level classification where the anomalous samples often differ significantly from the normal data either in semantic or visual. 
For instance, if we take images of cats as normal, then all the other images differ visually will be detected as anomalies, such as images of dogs. As recently Bergmann et al.~\cite{mvtec, ST}~suggested, fewer efforts have been done to improve algorithms for detecting anomalies that deviate subtly from the normal data.
Defects within images, as~Fig.\ref{multi-level}~shows, typically belong to this sort of anomaly. They often appear in very small and confined local regions of images and result in subtle visual deviations from the whole. In this scenario, an image-level anomaly detection algorithm may not be competent, especially when expecting to segment out the anomaly. 
One strategy to address the problem above is to exploit convolutional autoencoders~\cite{mvtec, ae-ssim-l2, ae-texture}. Generally, it first trains an autoencoder on anomaly-free images and then can perform pixel-precise anomaly detection during inference by comparing the pixel-wise differences between the input and its reconstructed version via some distance metric, e.g.~$l_2$-distance~\cite{l2}~or structural similarity metric (SSIM)~\cite{ssim}. 
Approaches based on this strategy assume that autoencoders solely trained on normal images are unable to reproduce the anomalous image subregions that deviate far from the normal ones. Thus, anomalies can be indicated by large reconstruction errors.
Deep generative models based on variational autoencoder (VAE)~\cite{vae}~and generative adversarial nets (GAN)~\cite{gan} could also used in a similar way~\cite{ae-ssim-l2,vae-l1,f-anogan}. Differently, generative methods can further leverage the reconstruction probability or likelihood score as an additional anomaly measurement~\cite{ae-ssim-l2,vae-prob,gan-avid,gan-prl}. 
These methods depended on image regeneration usually struggle to reproduce the sharp edges and complex textural structures. As a consequence, they always get high reconstruction errors in those edge or texture areas of images and incur many false anomaly alarms.
There are also detection approaches that leverage discriminative image features. They model the normality as well as inference the anomaly in the feature space. Typically, this category of method firstly divides an image into many partially overlapping regions or patches and construct corresponding region-level representations with either handcraft features~\cite{texem-gmm,texem-gmm-cs,atom-fd,multi-atom-fd}~or learned embeddings with convolutional neural networks~\cite{cs-fd,cnn-fd,ST}. Then, with the obtained regional features of the normal images, many machine learning methods can be used to model the distribution of normal feature patterns, such as gaussian mixture models~\cite{texem-gmm,texem-gmm-cs}, sparse coding~\cite{atom-fd,multi-atom-fd,cs-fd} and kmeans clustering~\cite{cnn-fd}. During inference, any regional feature pattern that deviates from the modeled distribution will be classified as an anomaly. Note that these feature-based methods accomplish anomaly detection and location simultaneously because the detection is done over every subregion of images. Since having to extract features for every local region of images, feature-based approaches are not very efficient especially when feature embeddings with deep neural networks are required~\cite{cnn-fd,ST}.
Besides, the spatial size of the region used for producing regional features affects the performance of feature-based models to a large extent.
If a smaller region size is selected, the extracted regional features may fail to capture large spatial structures within the image and also be sensitive to local changes of the image content, thus tending to either miss or wrongly report anomalies. 
In turn, if a larger size is used, the features may predominantly describe anomaly-free characteristics and ignore the traits of small anomalous regions at all.
A general practice to tackle the problem is by multi-scale modeling~\cite{texem-gmm, multi-atom-fd, ae-texture, ST}.
Multiple scale can be realized either by tweaking the scales of the input image~\cite{texem-gmm, multi-atom-fd} or controlling the sizes of the receptive field of convolutional neural networks~\cite{ae-texture, ST}.
At each scale, an anomaly detection model is trained and tested separately. To obtain the final multi-scale detection result, one has to combine all the detection results from different scales together via some rule, such as weighted averaging.
%
%
%
Indeed, the multi-scale strategy usually contributes to improved performance. But it also costs much more time at both training and testing stages.
In this work, we also propose to leverage the idea of multi-scale modeling. However, we propose to make full use of the pre-trained deep convolutional neural networks (CNNs). On the one hand, deep CNNs pre-trained on large datasets, such as ImageNet~\cite{imagenet}, can produce very discriminative features that have successfully transferred to many supervised vision tasks, such as edge detection~\cite{edge-hed, edge-rcf}, semantic segmentation~\cite{seg_fcn, semantic-seg}, as well as unsupervised anomaly detection~\cite{fcn-video-anomaly, occ-tf,occ-tf-m,ST}. On the other hand, with the deep hierarchical convolution architecture, the feature representations learned by different convolutional layers are inherently multi-scale~\cite{edge-multi-scale, edge-hed}. Each feature map, i.e. the output of each convolutional layer, is derived from a specific receptive field, and each location on the feature map perceives a corresponding spatial region of the input image. Therefore, each feature map in fact forms a dense regional representation for the whole image~\cite{seg_fcn, fcn-video-anomaly}, where each feature on the map represents a corresponding local region within the image, and the spatial size or scale of this region corresponds to the size of the specific receptive field.
Therefore, if, in a way, we fuse these hierarchical CNN feature maps, then a multi-scale dense regional feature representation of an image will be obtained by nature.

Based on the observations above, we specifically develop a regional feature generator that can align and aggregate the output feature maps of different convolution layers from a pre-trained deep CNN and produce multi-scale discriminative representations for every subregion of the input image via only one forward pass through the deep network. Since these regional features are very descriptive and can be generated efficiently, they are particularly beneficial for the task of unsupervised anomaly detection.
Besides, to leverage these dense regional features for effective and fast detection, we specifically design a deep yet efficient convolutional autoencoder (CAE) and detect possible anomalous regions within images through fast compressing and regenerating the dense regional representation. 
We term our anomaly detection method as Deep Feature Reconstruction (DFR), where we realize unsupervised anomaly detection and localization by reproducing the dense regional features generated from deep pre-trained CNNs with a deep CAE. Extensive experiments have been carried out to demonstrate that our DFR is both effective and computationally efficient.

\section{Related Work}
In the literature, the approaches specifically developed for unsupervised anomaly segmentation can be roughly divided into two categories: reconstruction-based and feature-based methods.

A typical group of reconstruction-based methods is based on convolutional autoencoders~\cite{mvtec,ae-texture, ae-ssim-l2}.
These models solely train on normal images and then detect anomalies within an image by computing the pixel-wise distances between the image and its reconstruction such as~$l_2$-distance~\cite{l2} and structural similarity metric (SSIM)~\cite{ssim}.
They assume that autoencoders trained on normal data are unable to reproduce the anomalous ones.
Deep generative models based on variational autoencoder (VAE)~\cite{vae}~and generative adversarial nets (GAN)~\cite{gan}~can also be used in similar ways. 
Baur et al.~\cite{vae-l1} utilize VAEGAN to detect anomalies or lesions in 2D Brain MR Images where the GAN component is used for adversarial training to enhance the reconstruction quality. During testing, they only use per-pixel~$l_1$-distance~to scoring the anomaly. Schlegl et al.~\cite{f-anogan} implement similar ideas for detecting anomalies in optical coherence tomography images but using a convolutional autoencoder instead.
Except the ordinary distance metrics, detection methods based on deep generative models can further leverage the reconstruction probability~\cite{ae-ssim-l2,vae-prob} and likelihood score~\cite{gan-avid,gan-prl}~as the additional anomaly measurements.
Besides, instead of comparing the differences between the input test image with its reconstruction, some methods propose to compute the residuals between the test image with its nearest normal counterpart. 
Schlegl et al.~\cite{anogan} propose AnoGAN where they train a GAN only on the normal images, and then detect anomalies by comparing differences between the test image and its nearest normal counterpart generated by the GAN. Specifically, they have to firstly search the nearest latent code of the test image in the GAN's latent space through an optimization process. With the obtained code, only then can they generate the expected nearest normal image for comparison.
David et al.~\cite{vae-grad}~also propose to detect the anomaly by comparing the differences between the test image and its nearest normal version. They train a VAE on the normal data then find the nearest normal counterpart for the test image by iterative updating the input of the VAE via gradient descent of a reconstruction loss defined on the test image and the VAE's output.
Both of the two methods need a searching step, thus they are not very efficient in the practice.
Since reconstruction-based methods detect anomalies within images in pixel or image space, they are usually required to produce high qualified images for comparison. However, the problem itself, i.e. high qualified image generation, is still challenging.

Unlike reconstruction-based models that detect anomalies in image space, feature-based methods detect anomalies in feature space. These approaches devote to construct descriptive representations for every local patch or region of the images with either handcraft features~\cite{texem-gmm,texem-gmm-cs,atom-fd,multi-atom-fd}~or embeddings produced by neural networks~\cite{cs-fd,cnn-fd,ST}.
Then relevant machine learning models, such as sparse coding~\cite{atom-fd,multi-atom-fd,cs-fd}, gaussian mixture models~\cite{texem-gmm,texem-gmm-cs} and kmeans clustering~\cite{cnn-fd}, are used to learn the distribution of the normal regional features. 
During inference, if a regional feature corresponding to a local region of the test image deviates from the learned distribution, then an anomalous region is detected. 
To further enhance the detection performance, multi-scale models are usually adopted~\cite{texem-gmm, multi-atom-fd}, where they will combine multiple models derived from different image region sizes together.


Recently, Bergmann et al.~\cite{mvtec}~developed a comprehensive benchmark dataset for unsupervised anomaly segmentation, which consists of various texture and object categories with over 70 different types of anomalies. They evaluated many state-of-the-art reconstruction-based and feature-based methods on this dataset, and found that none of these approaches work consistently well and a considerable improvement room exists.
More recently, Bergmann et al.~\cite{ST}~have proposed a novel unsupervised anomaly segmentation approach based on the student–teacher framework and achieved much better results than previous methods on the MVTec AD dataset. 
They leverage the transferred deep CNN features and detected anomalies in images via feature regression. 
Specifically, they train a knowledgeable teacher network on a large dataset with the guidance of pre-trained deep CNNs (e.g. resnet18~\cite{resnet}), and a group of student networks that imitates the teacher's behaviors solely on the anomaly-free data.
During testing, the students are utilized to predict the teacher's output, and anomaly scores are computed based on the corresponding predicting errors and uncertainties. 
The assumption lies that the student only trained to regress the teacher's output on the normal image patches well will probably predict poorly or fail to follow the teacher on the anomalous ones.
Besides, the authors also suggest using multi-scale models to enhance the final detection performance, i.e. an ensemble of multiple student-teacher pairs with different image patch sizes or various receptive fields.
In our work, we also propose to leverage the transferred deep CNN features and especially the multi-scale modeling. However, we suggest to build a multi-scale feature representation instead of the model ensemble and detect anomalies via feature reconstruction.

%
%
%

\section{Method}
\begin{figure*}
	\centering
	\includegraphics[width=1.\linewidth]{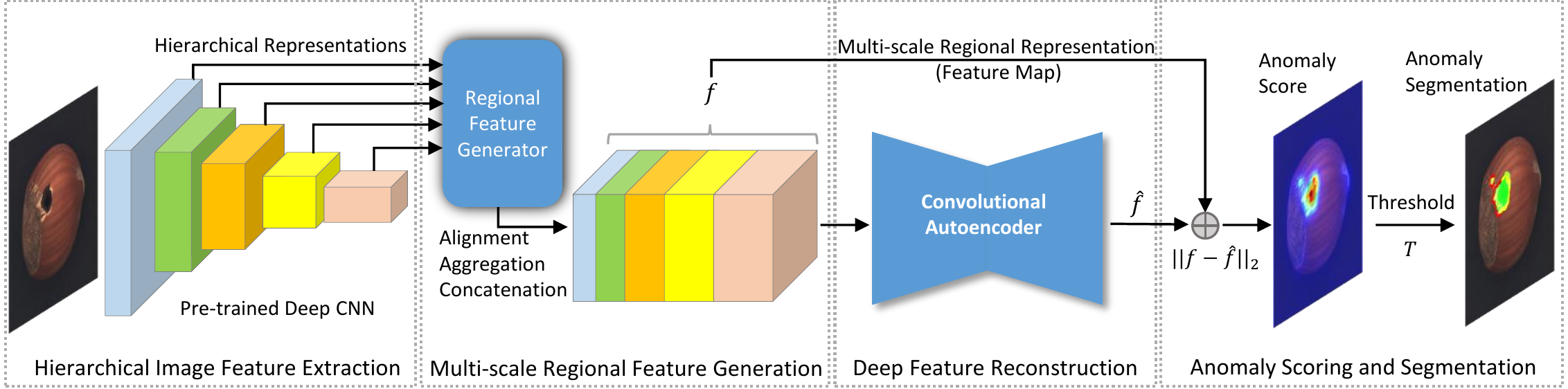}
	\caption{The overview of our unsupervised anomaly segmentation pipeline which consists four stages: hierarchical image feature generation, multi-scale regional feature generation, deep feature reconstruction, and anomaly scoring and segmentation. Figure Best viewed in color.}
	\label{framework}
\end{figure*}
The pipeline of our approach for unsupervised anomaly segmentation is outlined in~Fig.\ref{framework}. 
It has four stages, i.e. hierarchical image feature extraction, multi-scale regional feature generation, deep feature reconstruction, and scoring and segmentation.
Given an input image, firstly, the discriminative hierarchical image features (feature maps) are extracted via a pre-trained deep CNN.
Then, a regional feature generator takes the hierarchical feature maps as input and transforms them into a single feature map of a relatively large volume, which in essence establishes a dense multi-scale regional representation for the whole input image (Detailed explanations are in subsection~$B$).
Followed, a deep CAE convolves over the multi-scale representation and attempts to reproduce it again. 
Finally, to detect and segment the anomalous regions within the image, the reconstruction error and anomaly score map are calculated. The anomalies are segmented out if any scores on the anomaly map are larger than an estimated or user-defined threshold. We detail our pipeline in the following subsections.



\subsection{Hierarchical Image Feature Generation}
We use a pre-trained CNN to generate rich hierarchical discriminative features for the input image and then feed them into the multi-scale regional feature generator.

Suppose there is a convolutional neural network with~$L$~convolutional layers, typically each of which implements a composition of functions such as Convolution, Batch Normalization (BN)~\cite{bn}~and Rectified Linear Units (ReLU)~\cite{relu}.
Let~$\bm{x}$~with height~$h$, width~$w$~and channel~$c$~be an image. Passing it through the network, we can obtain a set of output feature maps~$\{\phi_{1}(\bm{x}), \phi_{2}(\bm{x}), ..., \phi_{L}(\bm{x})\}$~from the~$L$~convolutional layers, where the~$l$th feature map is of size~$h_{l}\times{w_{l}}\times{c_{l}}$.

Since each feature map is derived from a network layer in a specific depth with a specific receptive field (which can perceive a corresponding spatial region of the image), it comprises a certain level of representation or abstraction for the input image~\cite{edge-hed, multi-level_feat}.
The shallow convolutional layers with relatively small receptive fields capture the low-level characteristics such as the textural structures within the image. With the layers going deeper and their receptive fields becoming larger, the corresponding output feature maps encode more global or higher-level information such as an object or object parts in the input image.
Therefore, an ensemble of the feature maps~$\{\phi_{l}(\bm{x})\}_{l=1}^{L}$~naturally forms a rich hierarchical representation of the input image from the local details to the global semantic information. As an example, we detail the numbered convolutional layers and corresponding receptive field (RF) sizes of the VGG19~\cite{vgg19}~in~TABLE~\ref{vgg19}. The VGG19 net consists of 16 convolutional layers, and its receptive field size grows gradually from 3 to 252 as the layer becomes deeper.
Thus, here, the VGG19 can produce 16 different levels of representations for an input image.
\begin{table}[]
	\centering
	\caption{The numbered convolutional layer and corresponding receptive field size of VGG19}
	\resizebox{1.\columnwidth}!{
		\begin{tabular}{ccccccccc}
			\hline
			\textbf{Layer}   & \textbf{1} & \textbf{2} & \textbf{3} & \textbf{4} & \textbf{5} & \textbf{6} & \textbf{7} & \textbf{8} \\ \hline
			RF size & 3        & 5      & 10     & 14     & 24     & 32     & 40     & 48     \\ \hline \hline
			\textbf{Layer}   & \textbf{9}   & \textbf{10} & \textbf{11} & \textbf{12} & \textbf{13} & \textbf{14} & \textbf{15} & \textbf{16} \\ \hline
			RF size & 68       & 84     & 100    & 116    & 156    & 188    & 220    & 252    \\ \hline
		\end{tabular}
	}
	\label{vgg19}
\end{table}

\subsection{Multi-scale Regional Feature Generation}
%
%
%
%
%
%
%

\begin{figure}
	\centering
	\includegraphics[width=1.\linewidth]{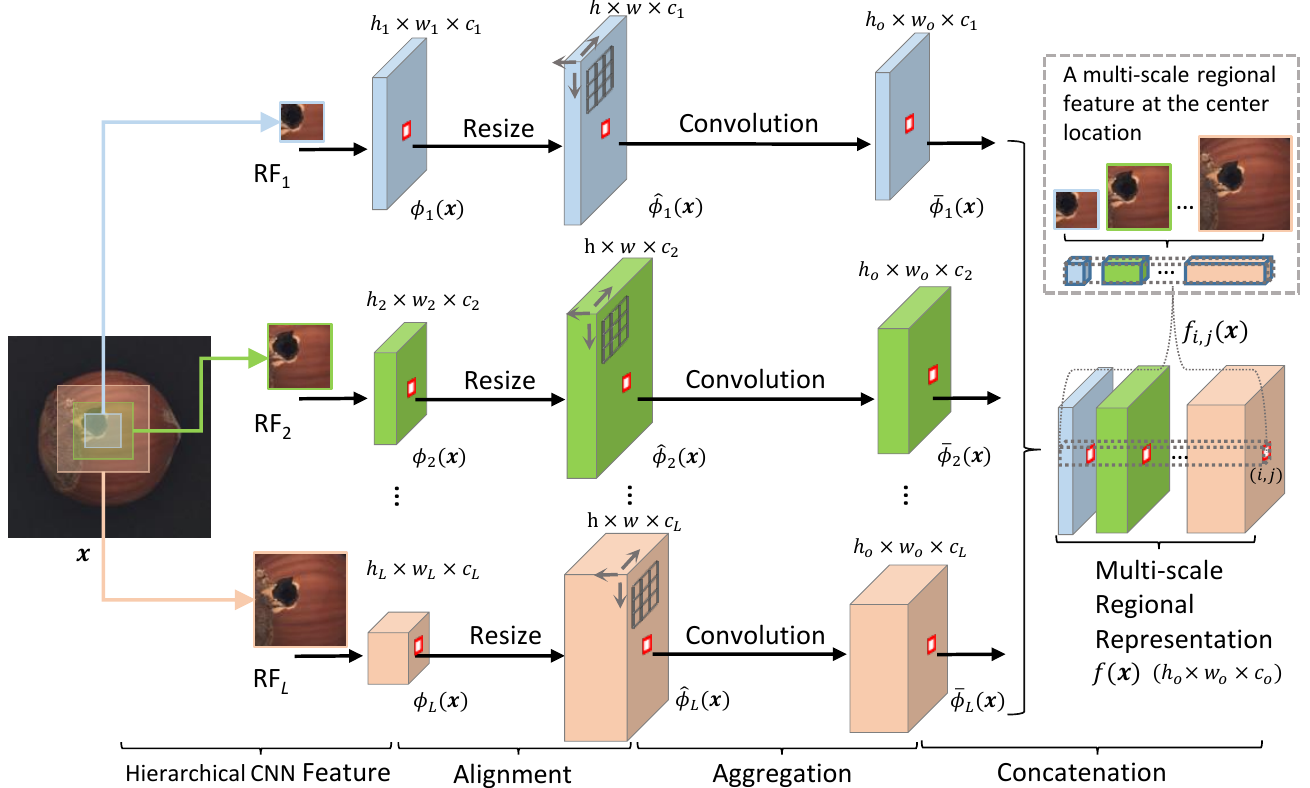}
	\caption{An illustration of the proposed multi-scale regional feature generator. Figure best viewed in color.}
	\label{regional-descriptor-generator}
\end{figure}

With the hierarchical CNN feature maps as input, we design a regional feature generator which can generate discriminative multi-scale representations for every subregion of the image. The overall scheme is shown in Fig.~\ref{regional-descriptor-generator}.

We firstly align the CNN feature maps~$\{\phi_{l}(\bm{x})\}_{l=1}^{L}$~that derived from different receptive fields by resizing all of them to the spatial size of the input image ($h\times{w}$) but with channels retrained:
\begin{equation}
	\hat{\phi_{l}}(\bm{x})=resize(\phi_{l}(\bm{x}))
\end{equation}
where the aligned~$l$th~feature map~$\hat{\phi_{l}}(\bm{x})$~has a size of~$h\times{w}\times{c_{l}}$.
Then a convolution operation is followed, where a mean filter is used to spatially convolve over the every aligned feature map with an appropriate stride. 
This is the aggregation operation:
\begin{equation}
\bar{\phi_{l}}(\bm{x})=agg(\hat{\phi_{l}}(\bm{x}))
\end{equation}
where the size of the~$l$th aggregated feature map is~$h_{o}\times{w_{o}}\times{c_{l}}$. The aggregation operation has two functions: first, it smooths the feature variations on the feature maps making the generated feature more robust to noisy input; second, it provides a way to control the spatial size of the aggregated feature representation such as by varying the convolution stride. 

Finally, we concatenate all the aggregated feature maps to a single feature map with the size of~$h_{o}\times{w_{o}}\times{c_{o}}$:
\begin{equation}
f_{\{1:L\}}(\bm{x})=cat(\bar{\phi_{1}}(\bm{x}),\bar{\phi_{2}}(\bm{x}),...,\bar{\phi_{L}}(\bm{x}))
\end{equation}
where~$f_{\{1:L\}}(\bm{x})$~denotes the resulted feature map combined from the 1 to~$L$th aggregated feature maps, and its depth or number of channels~$c_{o}$~is such that ~$c_{o}=\sum_{l=1}^{L}c_{l}$. For convenience, in some places, we also take~$f(\bm{x})$~instead of~$f_{\{1:L\}}(\bm{x})$~for short in the rest of the paper.

Obviously, the obtained final feature map is in fact a fusion of a series of transformed hierarchical CNN feature maps.
If we take a closer look, the fused feature map, in essence, forms a dense multi-scale regional description for the input image.
As Fig.\ref{regional-descriptor-generator} illustrates, each branch CNN feature map is derived from a convolutional layer with a specific receptive field. Each feature on a certain branch feature map describes an image subregion of a specific spatial size that equals the corresponding receptive field  (Note that, in Fig.\ref{regional-descriptor-generator}, we have only visualized what the feature on the center location can perceive from the image). When transforming all the hierarchical or branch CNN feature maps with operations such as alignment, aggregation, and concatenation, into a single feature map of a large volume, we naturally get a dense multi-scale representation for every local region of the image.
It is dense and multi-scale because every multi-scale feature~$f_{i,j}(\bm{x})$~with a dimension of~$c_{o}$~on the obtained feature map~$f(\bm{x})$~comprises a multi-scale description for a corresponding subregion on the image, where~$(i,j)$~denotes a spatial location on this feature map. In particular, each feature corresponds a non-overlapping region of the image with a spatial size of~$h/h_{o}\times{w/w_o}$. For instance, if the image is of size~$256\times{256}$ and the feature map is of spatial size~$64\times{64}$, then a feature on the map represents a~$4\times{4}$ pixel region of the image. 
If we make the final feature representation~$f(\bm{x})$~the same spatial size as the image, i.e.~$h=h_{o}$~and~$w=w_o$, then a pixel-wise representation will be obtained.

Note that though we use a multi-scale regional feature to represent a corresponding local region of the image, we usually derive the feature at each scale from a larger region or receptive filed. 
Such a multi-scale regional feature not only describes the local characteristics of the subregion itself but also encodes its multiple spatial context information or global characteristics, making it discriminative and very beneficial for anomaly detection.


In addition, we define that the scales of our regional representation correspond to the hierarchical layers of the CNN, and the spatial size of a specific scale equals the receptive field size of the corresponding convolutional layer.
For example, if using all the hierarchical convolutional layers of the VGG19, we will finally get a multi-scale regional representation with 16 different scales. And the size of each scale that corresponds to the specific receptive field of each convolutional layer can be found in TABLE~\ref{vgg19}. 
Moreover, one can also flexibly select different combinations of feature scales to meet the application requirement.

\subsection{Deep Feature Reconstruction}
The multi-scale regional features are discriminative. However, the feature dimension, i.e.~$c_o$, is usually very large.
To leverage such high-dimension regional representations for effective and fast anomaly detection, we design an efficient convolutional autoencoder which only includes operations of~$1\times{1}$~convolution and ReLU activation. Specifically, we use the CAE to convolve over the dense multi-scale regional representation~$f(\bm{x})$~and compress it into a low-dimension latent space, then manage to reproduce the representation again. The input representation~$f(\bm{x})$~and it reconstruction~$\hat{f}(\bm{x})$ will be use to score and segment the anomaly at the next stage of our pipeline.




We train the CAE solely on the regional representations of normal images with a reconstruction loss measured by the averaged pair-wise~$l^2$-distance between the reproduced dense regional representation~$\hat{f}(\bm{x})$~and its ground truth~$f(\bm{x})$:
\begin{equation}
	\mathcal{L}_{rec}=\sum_{i=1}^{h_o}\sum_{j=1}^{w_o}||f_{i,j}(\bm{x})-\hat{f}_{i,j}(\bm{x})||_2
\end{equation}

Note that both~$f(\bm{x})$~and~$\hat{f}(\bm{x})$~are in fact feature maps with the same size of~$h_{o}\times{w_o}\times{c_{o}}$, and each regional feature~$f_{i,j}(\bm{x})$~with dimension~$c_o$~on the regional feature map~$f(\bm{x})$~corresponds a local region of the input image. The detailed architecture of our CAE is in Appendix \textit{B}.




%

\subsection{Anomaly Scoring and Segmentation}
At the last stage of our pipeline, we detect all the possible anomalous regions within the input image base on the reproduced regional feature map~$\hat{f}(\bm{x})$~and its ground truth~$f(\bm{x})$. 
We first inference the anomaly score map by comparison of the ground truth representation~$f(\bm{x})$~and its reconstruction~$\hat{f}(\bm{x})$ and then binarize the anomaly map with a certain threshold to segment the anomaly.

We define the anomaly score map or anomaly map as the pair-wise reconstruction error between the input regional feature map~$f(\bm{x})$~and its reconstruction~$\hat{f}(\bm{x})$:
\begin{equation}
A_{i,j}(\bm{x})=||f_{i,j}(\bm{x})-\hat{f}_{i,j}(\bm{x})||_2
\end{equation}
where~$A_{i,j}(\bm{x})$~is the anomaly score of the regional feature~$f_{i,j}(\bm{x})$, and~$(i,j)$~denotes the spatial location where the regional feature~$f_{i,j}(\bm{x})$ lies on the regional feature map~$f(\bm{x})$ of the input image~$\bm{x}$. 
Correspondingly,~$A(\bm{x})$~is the regional anomaly map of the image with the same spatial size as~$f(\bm{x})$, i.e.~$h_{o}\times{w_o}$. To obtain a pixel-wise anomaly map~$\hat{A}(\bm{x})$~for the image, we further bilinearly upsample the regional anomaly map to the same spatial size of the image.

We assume that the CAE solely trained on the regional features of normal images are unable to reproduce the regional features correspond to anomalous image regions. Therefore, anomalous regions coincide with the large reconstruction errors of the corresponding regional features or the high scores on the anomaly map.

To get the final segmentation result, we binarize the anomaly map~$\hat{A}(\bm{x})$ with a threshold~$T$. 
Specifically, we use the acceptable false positive rate (FPR) on the normal data to estimate the segmentation threshold. For instance, if the acceptable FPR is expected to be zero, then it means that the threshold should be such that no pixels in the normal images are wrongly classified as anomalies. If the FPR is $0.005$, then the segmentation threshold should meet that just $0.5$ percent of pixels in the normal images are incorrectly detected as anomalous.

	
\section{Experiments}
In this section, we first present the experimental comparisons with state-of-the-art unsupervised anomaly segmentation methods and then conduct a thorough analysis of our approach. 

\subsection{Experimental Setup}
\subsubsection{Datasets}
We evaluate the proposed approach on the challenging MVTec Anomaly Detection (MVTec AD) dataset~\cite{mvtec}, which is specifically developed to benchmark unsupervised algorithms for anomaly segmentation. It includes a collection of 15 sub-datasets (10 for objects and 5 for textures) and contains a total of 5354 high-resolution images with over 70 types of anomalies such as scratches, cracks, stains, and various structural damages. 
All the datasets are divided into training and testing sets, where the training sets are only consist of normal images while the testing sets contains both normal and anomalous samples.
Detailed statistics of the MVTec AD dataset is in Appendix~$A$.
\subsubsection{Baselines}
We compare our approach against the following state-of-the-art unsupervised anomaly segmentation methods:
\begin{itemize}
	\item AE-$l_2$ and AE-$ssim$~\cite{ae-ssim-l2}: approaches based on CAEs, which detect anomalies by pixel-wise comparisons between the input image and its reconstruction via $l_2$-distance~\cite{l2}~and SSIM~\cite{ssim}~respectively.
	\item AnoGAN~\cite{anogan}: a GAN based model, which first tries to generates the nearest normal image for the test image with the GAN generator trained only on the normal data and then detects the anomaly by computing per-pixel residuals between the test and its nearest normal counterpart.
	\item VAE-$grad$~\cite{vae-grad}: a recently proposed reconstruction-based method, which first trains a VAE solely on normal data, then attempts to find the nearest normal image for the test image by iterative updating the VAE's input via minimizing a reconstruction loss defined on the test image and the VAE's output, finally detects the anomaly by comparing differences between the test image and its nearest normal version, i.e. the VAE’s input that obtained at the final iteration.
	\item CNN-FD~\cite{cnn-fd}: a method which exploits deep CNN features and uses a shallow model, i.e. kmeans clustering, to learn the normality and inference the anomaly.
	\item ST~\cite{ST}: a recently proposed powerful anomaly segmentation approach, which leverages both deep CNN embeddings and multi-scale modeling, and detects anomalous regions within images using a student–teacher framework. Specifically, we will compare our method with two best performing models in~\cite{ST}, i.e. the ST-m and ST-p65, where ST-m is a multi-scale ST model and ST-p65 a single scale model established on image patches of size~$65\times{65}$.
\end{itemize}

\subsubsection{Architecture Details}
We take the VGG19~\cite{vgg19}~pre-trained on ImageNet~\cite{imagenet}~to produce hierarchical image features. In particular, we strip the last 3 dense layers and only retain the front 16 convolutional layers. We get CNN feature maps from the ReLU outputs of the convolutional layers and number the feature maps with the order of the corresponding layer in the network. 
The ordered layer and corresponding receptive field are listed in TABLE~\ref{vgg19}.
Since we have defined that the scales and scale sizes of our regional features correspond to the hierarchical convolutional layers and their receptive fields respectively, we will get at most 16 different scales of regional features with the trimmed VGG19.
For our regional feature generator, we align the hierarchical CNN feature maps by nearest-neighbor interpolation and use a mean filter with spatial size of~$4\times{4}$~and stride of 4 to aggregate the aligned hierarchical CNN feature maps. As a result, we obtain such a regional feature representation where each location on the regional feature map corresponds to a~$4\times{4}$ pixel region of the input image. 
As for the CAE, we first randomly sample a subset of regional features from the regional feature map, then estimate the latent code dimension with Principal Component Analysis (PCA) such that $90\%$ variance is just explained. The concrete parameters of our CAE depends on the dataset and the number of CNN feature maps that used. The CAE architecture is detailed in Appendix $B$. 

 
\subsubsection{Training Details}
For all experiments, the images are resized to the size of~$256\times{256}$ pixels and their channels are triplicated if gray images are encountered.
For all the datasets, we train our model solely on the anomaly-free training sets using the Adam optimizer with a learning rate of~$1\times{10^{-4}}$~and a batch size of 4 for 700 epochs.
During training, we freeze the weights of the pretrained VGG19 and the regional feature generator, and only update the weights of the CAE.
We implements our method in Pytorch with a NVIDIA GeForce GTX 1080 Ti. The code is publicly available$\footnote{https://github.com/YoungGod/DFR}$.

\subsubsection{Evaluation Metrics}



We take the area under the receiver operating characteristic curve (ROC-AUC)~\cite{mvtec, vae-grad}~and the area under the per-region-overlap curve (PRO-AUC)~\cite{mvtec, ST}~as our evaluation metrics. The ROC-AUC assesses the best potential segmentation result in terms of normal and anomalous pixels, i.e. per pixel overlapping performance.
The PRO-AUC metric is suggested in~\cite{mvtec, ST}. It attempts to measure the best possible segmentation performance across normal and anomalous regions at the region level, i.e. per region overlapping performance.
Specifically, the PRO-AUC weights all the ground-truth anomalous regions equally so that the segmentation performance is measured with no bias to either large or small ground-truth regions.
In other words, it measures a model's ability to segment out all the possible anomalous regions equally no matter what the size of a particular abnormal region is. 
Simple per-pixel segmentation metrics such as PRO-AUC may fail to measure this property since a large enough correctly segmented region can compensate for many wrongly segmented minor ones.
As~\cite{ST}~suggests, we report the normalized PRO-AUC up to an average per-pixel false positive rate (FPR) of 30\%, where the average FPR is the percentage of pixels within all testing images that are incorrectly detected as anomalies. We calculate the PRO-AUC metric as in~\cite{ST}.









\subsection{Comparisons Against Baselines}
We evaluate the segmentation performances of our approach and baselines on the testing sets across 7 object and 5 texture data categories. For comparison, the ROC-AUC and PRO-AUC metrics of our method are calculated under two settings: 1) all the 16 scales of the regional representation are used, i.e.~$f_{\{1:16\}}$; 2) only the front of 12 scales are taken, i.e.~$f_{\{1:12\}}$. Besides, we take the ROC-AUC results of baselines from~\cite{mvtec}~and~\cite{vae-grad}, and PRO-AUC results from~\cite{ST}~respectively.

TABLE~\ref{table-auc}~shows the ROC-AUC results. Our methods outperform the baselines on most of the data categories, except \textit{Tile} and \textit{Transistor}. On average, ours improve the ROC-AUC performances of the baseline methods by a very large margin. 
TABLE~\ref{table-overlap}~is the PRO-AUC results. Comparing with AE-$ssim$, AE-$l_2$ and CNN-FD, our approaches achieve overwhelming results across all data categories. Besides, our method shows similar or better results on many data categories when compared with ST models, i.e. ST-m, ST-p65. Averaging the metrics over all categories, ours work on par with ST-m.


However, the ST model needs to train several different networks simultaneously, i.e. a teacher network and an ensemble of student networks. During inference, the test image has to be separately passed through the students and teacher networks to generate corresponding dense embeddings and compute the anomaly scores.
To make a multi-scale model, ST has to independently train multiple such pairs of student-teacher networks and then combine the separately detected results together during testing. 
While our approach is more direct yet effective, where a multi-scale anomaly detection can be realized with only one forward pass through our pipeline network. That is our method in fact is inherently multi-scale. Besides, we can also flexibly combine different scales to meet specific applications. The only part of our pipeline needing to train is the CAE.

It is also interesting to note that our methods, ST-m and ST-p65 work much better than the reconstruction-based models such as AE-$ssim$ and AnoGAN.
This is likely because that both ST models and ours leverage the transferred discriminative CNN features.
This also indicates that, for anomaly detection, approaches which leverage transferred discriminative features may show more potential than the methods which only learn representations from scratch such as models based on autoencoders and GANs. Similar findings are presented in~\cite{ST} and~\cite{dis-gan}.
Besides, CNN-FD also use the transferred deep features, but it shows inferior performance on most data categories. Since CNN-FD adopts a shallow algorithm, i.e. kmeans clustering, for anomaly detection, the shallow model is not capable to make full use of the rich CNN features due to its limited model capacity.
 
In addition, multi-scale modeling contributes to improved performance. As the results in TABLE~\ref{table-overlap} show, the multi-scale model ST-m works better than ST-p65 on average. 
And our approach achieves better average performance either on ROC-AUC or PRO-AUC metric when all the 16 scales of the regional features are used. These results indicate that each feature scale may conveys some useful information for anomaly detection. Thus, if there is no prior knowledge of the anomalies, a multi-scale model is usually a wise choice.

\begin{figure*}
	\centering
	\includegraphics[width=1.\linewidth]{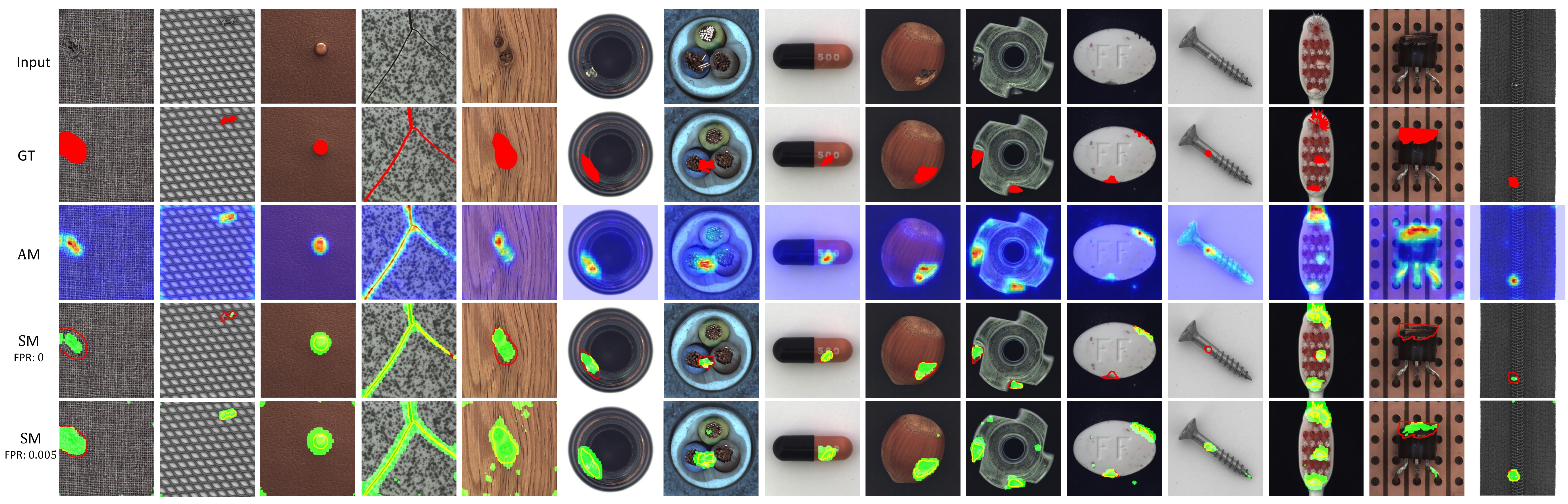}
	\caption{Qualitative results of our unsupervised anomaly segmentation approach. Input: input anomalous image. GT: ground truth anomalous regions (in red). AM: anomaly map (red regions correspond to high score for anomaly). SM: segmentation map. Note that the segmentation maps are visualized when acceptable FPRs of 0 and 0.005 on corresponding training data are given. Figure best viewed in color.}
	\label{seg-quality-l12}
\end{figure*}

We have also visualized some qualitative segmentation results over all the data categories in~Fig.~\ref{seg-quality-l12}. Specifically, the figure shows the obtained anomaly maps and segmentation maps of our approach when the front 12 scales of the regional representation,~i.e.~$f_{\{1:12\}}$, are used. For visualization, the anomaly maps are respectively normalized to the range~$[0,1]$ and superimposed on the corresponding testing images.
\begin{table}[]
	\centering
	\caption{Quantitative comparisons with baselines (ROC-AUC)}
	\resizebox{1.\columnwidth}!{
	\begin{tabular}{ccccccccc}
		\hline
		& Category   
		&\begin{tabular}[c]{@{}c@{}}AE\\$ssim$\end{tabular} 
		&\begin{tabular}[c]{@{}c@{}}AE\\$l_{2}$\end{tabular}
		&\begin{tabular}[c]{@{}c@{}}Ano-\\GAN\end{tabular} 
		&\begin{tabular}[c]{@{}c@{}}VAE\\$grad$\end{tabular}
		&\begin{tabular}[c]{@{}c@{}}CNN\\FD\end{tabular} 
		&\begin{tabular}[c]{@{}c@{}}Ours\\$f_{\{1:12\}}$\end{tabular}   
		&\begin{tabular}[c]{@{}c@{}}Ours\\$f_{\{1:16\}}$\end{tabular}   
		\\ \hline
		\multirow{5}{*}{\rotatebox{90}{Textures}} & Carpet     & 0.87         & 0.59   & 0.54   & 0.74     & 0.72          & 0.96          & \textbf{0.97} \\ \cline{2-9} 
		& Grid       & 0.94         & 0.90    & 0.58   & 0.96     & 0.59          & \textbf{0.98} & \textbf{0.98} \\ \cline{2-9} 
		& Leather    & 0.78         & 0.75   & 0.64   & 0.93     & 0.87          & \textbf{0.99} & 0.98          \\ \cline{2-9} 
		& Tile       & 0.59         & 0.51   & 0.50    & 0.65     & \textbf{0.93} & 0.86          & 0.87          \\ \cline{2-9} 
		& Wood       & 0.73         & 0.73   & 0.62   & 0.84     & 0.91          & \textbf{0.94} & 0.93          \\ \hline
		\multirow{10}{*}{\rotatebox{90}{Objects}} & Bottle     & 0.93         & 0.86   & 0.86   & 0.92     & 0.78          & 0.95          & \textbf{0.97} \\ \cline{2-9} 
		& Cable      & 0.82         & 0.86   & 0.78   & 0.91     & 0.79          & 0.88          & \textbf{0.92} \\ \cline{2-9} 
		& Capsule    & 0.94         & 0.88   & 0.84   & 0.92     & 0.84          & 0.98          & \textbf{0.99} \\ \cline{2-9} 
		& Hazelnut   & 0.97         & 0.95   & 0.87   & 0.98     & 0.72          & 0.98          & \textbf{0.99} \\ \cline{2-9} 
		& Meta Nut   & 0.89         & 0.86   & 0.76   & 0.91     & 0.82          & 0.90          & \textbf{0.93} \\ \cline{2-9} 
		& Pill       & 0.91         & 0.85   & 0.87   & 0.93     & 0.68          & 0.96          & \textbf{0.97} \\ \cline{2-9} 
		& Screw      & 0.96         & 0.96   & 0.80    & 0.95     & 0.87          & \textbf{0.99} & \textbf{0.99} \\ \cline{2-9} 
		& Toothbrush & 0.82         & 0.93   & 0.90    & 0.98     & 0.77          & 0.98          & \textbf{0.99} \\ \cline{2-9} 
		& Transistor & 0.90 & 0.86   & 0.80    & \textbf{0.92}     & 0.66          & 0.75          & 0.80          \\ \cline{2-9} 
		& Zipper     & 0.88         & 0.77   & 0.78   & 0.87     & 0.76          & \textbf{0.96} & \textbf{0.96} \\ \hline \hline
		& Mean       & 0.86         & 0.82   & 0.74   & 0.89     & 0.78          & 0.94          & \textbf{0.95} \\ \hline
	\end{tabular}}
\label{table-auc}
\end{table}

\begin{table}[]
	\centering
	\caption{Quantitative comparisons with baselines (PRO-AUC).}
	\resizebox{1.\columnwidth}!{
	\begin{tabular}{lllllllll}
		\hline
		& Category
		&\begin{tabular}[c]{@{}c@{}}AE\\$ssim$\end{tabular}
		&\begin{tabular}[c]{@{}c@{}}Ano-\\GAN\end{tabular} 
		&\begin{tabular}[c]{@{}c@{}}CNN\\FD\end{tabular} 
		&\begin{tabular}[c]{@{}c@{}}ST\\$p65$\end{tabular}      
		&\begin{tabular}[c]{@{}c@{}}ST-m\end{tabular}         
		&\begin{tabular}[c]{@{}c@{}}Ours\\$f_{\{1:12\}}$\end{tabular}   
		&\begin{tabular}[c]{@{}c@{}}Ours\\$f_{\{1:16\}}$\end{tabular}   
		\\ \hline
		\multirow{5}{*}{\rotatebox{90}{Textures}} & Carpet     & 0.65     & 0.20   & 0.47                   & 0.70          & 0.88          & \textbf{0.93} & \textbf{0.93} \\ \cline{2-9} 
		& Grid       & 0.85     & 0.23   & 0.18                   & 0.82          & \textbf{0.95} & 0.93          & 0.93          \\ \cline{2-9} 
		& Leather    & 0.56     & 0.38   & 0.64                   & 0.82          & 0.95          & \textbf{0.97} & \textbf{0.97} \\ \cline{2-9} 
		& Tile       & 0.18     & 0.18   & 0.80                   & 0.91          & \textbf{0.95} & 0.79          & 0.79          \\ \cline{2-9} 
		& Wood       & 0.61     & 0.39   & 0.62                   & 0.73          & 0.91          & \textbf{0.93} & 0.91          \\ \hline
		\multirow{10}{*}{\rotatebox{90}{Objects}} & Bottle     & 0.83     & 0.62   & 0.74                   & 0.92          & \textbf{0.93} & 0.92          & \textbf{0.93} \\ \cline{2-9} 
		& Cable      & 0.48     & 0.38   & 0.56                   & \textbf{0.87} & 0.82          & 0.77          & 0.81          \\ \cline{2-9} 
		& Capsule    & 0.86     & 0.31   & 0.31                   & 0.92          & \textbf{0.97} & 0.96          & \textbf{0.97} \\ \cline{2-9} 
		& Hazelnut   & 0.92     & 0.70   & 0.84                   & 0.94          & \textbf{0.97} & \textbf{0.97} & \textbf{0.97} \\ \cline{2-9} 
		& Meta Nut   & 0.60     & 0.32   & 0.36                   & 0.90          & \textbf{0.94} & 0.87          & 0.90          \\ \cline{2-9} 
		& Pill       & 0.83     & 0.78   & 0.46                   & 0.94          & \textbf{0.96} & \textbf{0.96} & \textbf{0.96} \\ \cline{2-9} 
		& Screw      & 0.89     & 0.47   & 0.28                   & 0.93          & 0.94          & 0.95          & \textbf{0.96} \\ \cline{2-9} 
		& Toothbrush & 0.78     & 0.75   & 0.15                   & 0.86          & \textbf{0.93} & \textbf{0.93} & \textbf{0.93} \\ \cline{2-9} 
		& Transistor & 0.73     & 0.55   & 0.63                   & 0.70          & 0.67          & 0.77          & \textbf{0.79} \\ \cline{2-9} 
		& Zipper     & 0.67     & 0.47   & 0.70                   & 0.99          & \textbf{0.95} & 0.89          & 0.90          \\ \hline \hline
		& Mean       & 0.69     & 0.45   & 0.52                   & 0.86          & \textbf{0.91} & 0.90          & \textbf{0.91} \\ \hline
	\end{tabular}}
\label{table-overlap}
\end{table}

\subsection{Analysis}
\subsubsection{Effectiveness of the multi-scale representation}
In the above experiments, to implement our approach, we respectively use two different multi-scale regional representations, i.e.~${f_{\{1:12\}}}$~and~${f_{\{1:16\}}}$. Both of them contains multiple feature scales, i.e. 12 and 16 respectively. 
However, is it reasonable to exploit so many scales? 
To answer this question, we conduct thorough experiments on all the data categories of objects and textures with different multi-scale regional representations.
Concretely, we evaluate our model in a series of scenarios where different multi-scale regional representations,~i.e.~$f_{\{1:2\}}$,~${f_{\{1:4\}}}$,~${f_{\{1:8\}}}$,~${f_{\{1:12\}}}$,~${f_{\{1:16\}}}$, are used respectively. 
Note that, from~$f_{\{1:2\}}$ to~${f_{\{1:16\}}}$, the number of different feature scales held by the corresponding multi-scale representation gradually increases from 2 to 16 and the corresponding maximum scale size goes from 5 to 252 (One can refer to TABLE~\ref{vgg19}~for these scale sizes), which means that more and larger scales are gradually exploited to form these different multi-scale representations.

\begin{figure}
	\centering
	\includegraphics[width=1.\linewidth]{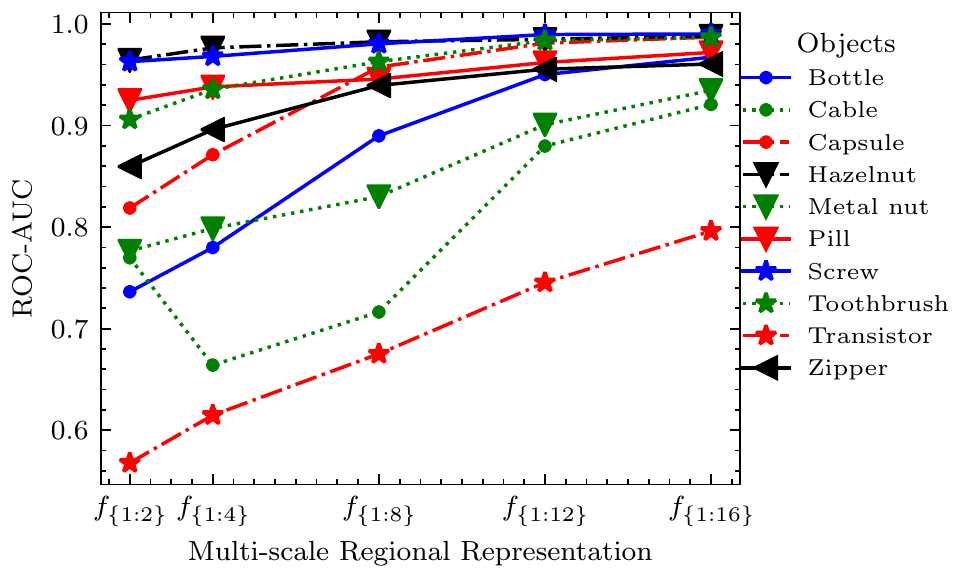}
	\caption{The ROC-AUC metrics of our approach on object categories with different multi-scale regional representations.}
	\label{auc_objects_s}
\end{figure}

\begin{figure}
	\centering
	\includegraphics[width=1.\linewidth]{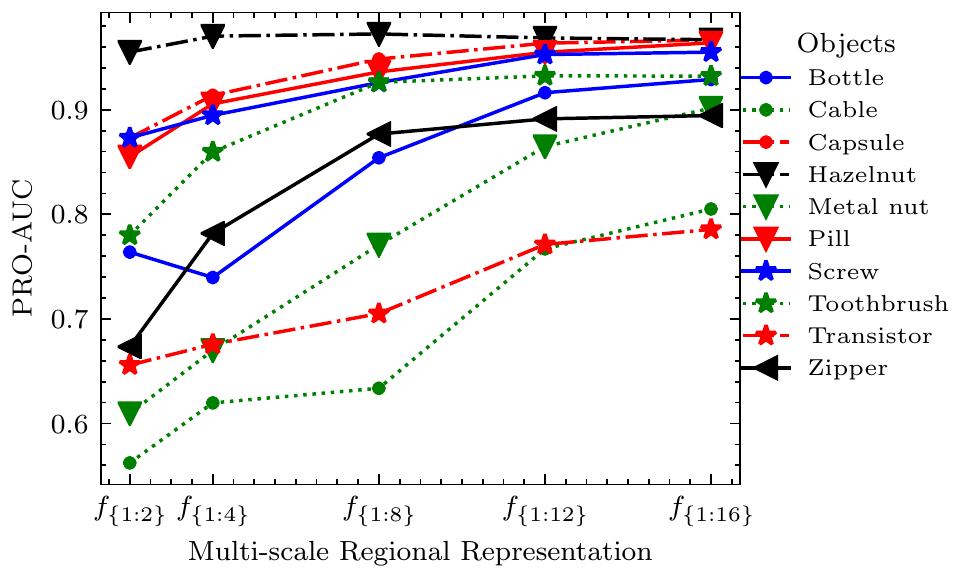}
	\caption{The PRO-AUC metrics of our approach on object categories with different multi-scale regional representations.}
	\label{overlap_objects_s}
\end{figure}

\begin{figure}
	\centering
	\includegraphics[width=.97\linewidth]{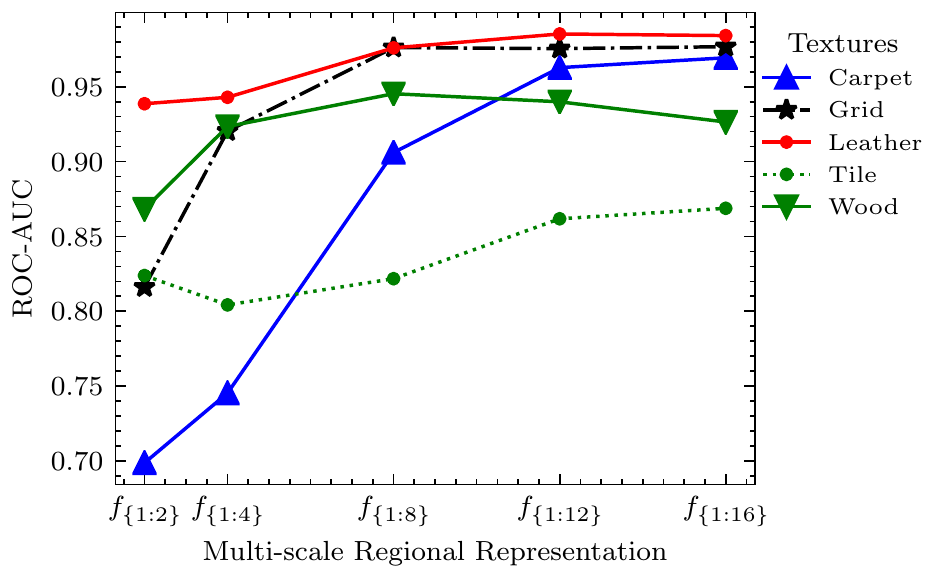}
	\caption{The ROC-AUC metrics of our approach on texture categories with different multi-scale regional representations.}
	\label{auc_textures_s}
\end{figure}

\begin{figure}
	\centering
	\includegraphics[width=.97\linewidth]{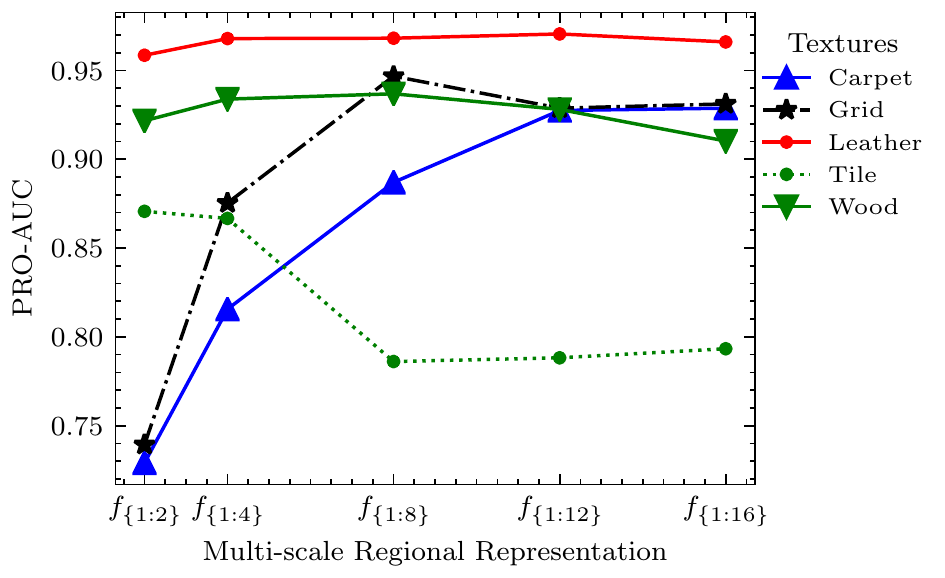}
	\caption{The PRO-AUC metrics of our approach on texture categories with different multi-scale regional representations.}
	\label{overlap_textures_s}
\end{figure}

The results for object data categories are shown in Fig.~\ref{auc_objects_s} and Fig.~\ref{overlap_objects_s}. 
In terms of ROC-AUC and PRO-AUC, all the object categories benefit from multi-scale representations. With more scales held by the multi-scale representation, the performance of our model becomes better. Similar phenomenon can be seen in Fig.~\ref{auc_textures_s} and Fig.~\ref{overlap_textures_s} for most of the texture categories, except \textit{Wood}. Though the ROC-AUC on~\textit{Wood} tends to increase when more scales are included in its multi-scale representation, the corresponding PRO-AUC decreases. This indicates that our model on \textit{Wood} prefers to detect larger anomalous regions when more and larger scales are used.
In addition, we can also observe that the metrics on textures saturate at~$f_{\{1:8\}}$ or~$f_{\{1:12\}}$ and even slightly degrade afterwards. 
This is because that relatively local statistics are usually enough to represent the textural structures.

Some qualitative results are also presented in Fig.~\ref{multi-level}. With more and larger scales taken into account, the corresponding anomaly maps tend to approach their ground truth counterparts progressively.
And the false anomalous regions are gradually removed, while the truth anomalous regions are gradually refined. 
This is because, with more scales leveraged, the corresponding multi-scale regional representation will encode more spatial context information for every subregion of an image, thus making the detection more certain or confident.  



%
Though we have demonstrated that our approach tends to perform better when leveraging more and larger scales, however, are the small scales still helpful? If not, we can drop them to build a more compact model.
To identify this, further experimental scenarios are designed. Concretely, we start from a regional representation derived from a large scale, i.e.~$f_{\{12\}}$, then gradually add the regional features from smaller scales to form a series of multi-scale representations,~i.e.~$f_{\{9:12\}}$,~${f_{\{5:12\}}}$,~${f_{\{1:12\}}}$.

The results are shown in TABLE~\ref{table_auc_overlap}. With more small scales considered, the average performances improve gradually but by a small margin. The results suggest that each scale of the regional representation conveys some different information that can advance the detection performance. In addition, the regional representations derived from relatively large scales may have contained much useful information for anomaly detection. As it can be seen from the TABLE~\ref{table_auc_overlap}, only with the regional representation~$f_{\{12\}}$, our model can also obtain satisfactory results.

In general, we can draw the following conclusions about our approach: 1) multi-scale modeling, i.e using multi-scale regional representation, is always beneficial for anomaly detection. With more and larger scales leveraged, the detection performance always tends to become better; 2) for texture categories, it is enough to exploit less and smaller feature scales when compared with the object categories; 3) we can drop some smaller scales to establish more compact models if not pursuing high detection metrics.

\begin{table}[]
	\centering
	\caption{Metrics with different multi-scale regional representations.}
	\resizebox{.9\columnwidth}!{
		\begin{tabular}{ccccccc}
			\hline
			Category                  & Metric  &$f_{\{12\}}$    &$f_{\{9:12\}}$ &$f_{\{5:12\}}$ &$f_{\{3:12\}}$ &$f_{\{1:12\}}$ \\ \hline
			\multirow{2}{*}{Textures} & ROC-AUC     & 0.922 & 0.938 & 0.940 & 0.941 & \textbf{0.945} \\ \cline{2-7} 
			& PRO-AUC & 0.877 & 0.897 & 0.903 & 0.906 & \textbf{0.909} \\ \hline
			\multirow{2}{*}{Objects}  & ROC-AUC     & 0.916 & 0.930 & 0.929 & 0.930 & \textbf{0.933} \\ \cline{2-7} 
			& PRO-AUC & 0.868 & 0.889 & 0.892 & 0.892 & \textbf{0.898} \\ \hline
	\end{tabular}}
	\label{table_auc_overlap}
\end{table}

\subsubsection{Boundary Effects}
\begin{figure}
	\centering
	\includegraphics[width=.8\linewidth]{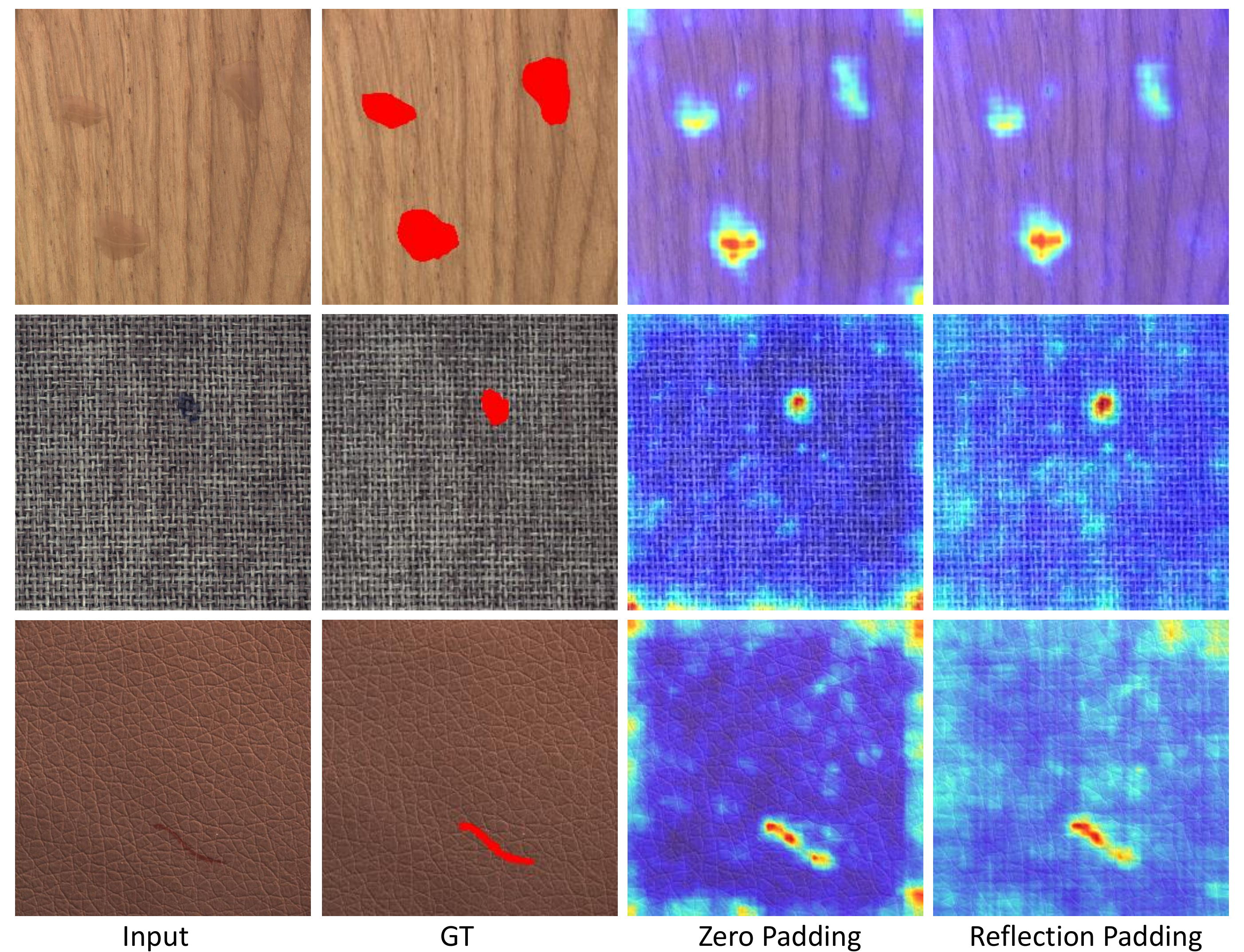}
	\caption{Examples of boundary effects. Input: input anomalous image. GT: ground truth anomalous regions (in red). The last two columns are respectively the resulted anomaly maps when using zero padding and refection padding. Note that red regions correspond to high score for anomaly. Figure best viewed in color.}
	\label{padding}
\end{figure}

We observe that our approach tends to wrongly report anomalies near the image boundary areas when the foreground (the target we are interested in) fills the whole image, such as the texture categories. One cause may be the zero-padding operation used in the pre-trained VGG19. 
This operation will inject novel information that is out of the image into the corresponding CNN feature maps, especially the boundary regions. Since these features in boundary areas are not statistically significant compared with the features in relatively center areas, our CAE may not model these feature patterns well. 
One possible solution is to adopt the reflection padding that only uses information from the image itself.
With reflection padding instead, we train and evaluate our model on all the data categories.  
As Fig.~\ref{padding} shows, the boundary effects are relieved with the reflection padding strategy.
In addition, as TABLE~\ref{table_padding}~presents, the averaged performances on both object and texture categories are improved by about 1 percent when the reflection padding is used. 

\begin{table}[]
	\centering
	\caption{Metrics with different padding modes.}
	\resizebox{0.65\columnwidth}!{
		\begin{tabular}{cccc}
			\hline
			Category                  & Meric   & Zero   & Refection       \\ \hline
			\multirow{2}{*}{Textures} & ROC-AUC     & 0.945 & \textbf{0.955} \\ \cline{2-4} 
			& PRO-AUC & 0.909 & \textbf{0.920} \\ \hline
			\multirow{2}{*}{Objects}  & ROC-AUC     & 0.933 & \textbf{0.941} \\ \cline{2-4} 
			& PRO-AUC & 0.898 & \textbf{0.899} \\ \hline
	\end{tabular}}
	\label{table_padding}
\end{table}

\subsubsection{Inference Speed}
We evaluate the inference speed of our method under many different multi-scale settings. Since the model architecture depends on the specific dataset, we average the inference time across all the object and texture categories on corresponding testing sets. TABLE~\ref{table-speed}~shows the statistics of the inference speed when using different multi-scale regional representations. We also list the corresponding average performances where reflection padding strategy is used. In general, our method can reach over about 100 frames per second (fps) even when 12 different feature scales are used, which indicates our approach is applicable in practice.
Besides, leveraging fewer feature scales, we can further obtain more compact and efficient models only with small degradations on performances.



\begin{table}[]
	\centering
	\caption{Inference speed and metric under different multiple scale settings.}
	\resizebox{0.85\columnwidth}!{
		\begin{tabular}{cccccc}
			\hline
			 &$f_{\{12\}}$    &$f_{\{9:12\}}$ &$f_{\{5:12\}}$ &$f_{\{3:12\}}$ &$f_{\{1:12\}}$ \\ \hline
			Speed (fps)   & 159 & 148    & 116    & 111    & 100    \\ \hline
			ROC-AUC   & 0.926  & 0.942    & 0.941    & 0.942    & 0.946    \\ \hline
			PRO-AUC   & 0.875 & 0.895    & 0.899    & 0.900    & 0.906    \\ \hline
	\end{tabular}}
	\label{table-speed}
\end{table}

\section{Conclusion}

In this work, we have primarily presented a general unsupervised approach, i.e. DFR, to detect anomalous regions within images. We propose to make use of the transferred hierarchical CNN features to build dense discriminative multi-scale feature representations for every local region of the images via a specially designed regional feature generator. We also propose to detect possible anomalous regions in images through deep feature reconstruction, i.e. reconstructing the multi-scale regional features via a deep yet efficient CAE. Extensive experiments and analysis on various data categories of objects and textures have demonstrated that our method is effective and achieves state-of-the-art results. 
In future work, we plan to further optimize our approach for more compact and efficient implementations.

\appendices
\section{MVTec AD dataset}
The detailed statistics of the MVTec AD dataset is given in~\ref{table-AD}.
\begin{table}[]
	\centering
	\caption{MVTec Ad dataset.}
	\resizebox{.85\columnwidth}!{
		\begin{tabular}{cccccc}
			\hline
			& Category   
			& Train 
			& Test 
			&\begin{tabular}[c]{@{}c@{}}Anomaly\\Types\end{tabular} 
			&\begin{tabular}[c]{@{}c@{}}Anomalous\\Regions\end{tabular}  \\ \hline
			\multirow{5}{*}{Textures} & Carpet     & 280   & 117  & 5             & 97                \\ \cline{2-6} 
			& Grid       & 264   & 78   & 5             & 170               \\ \cline{2-6} 
			& Leather    & 245   & 124  & 5             & 99                \\ \cline{2-6} 
			& Tile       & 230   & 117  & 5             & 86                \\ \cline{2-6} 
			& Wood       & 247   & 79   & 5             & 168               \\ \hline
			\multirow{10}{*}{Objects} & Bottle     & 209   & 83   & 3             & 68                \\ \cline{2-6} 
			& Cable      & 224   & 150  & 8             & 151               \\ \cline{2-6} 
			& Capsule    & 219   & 132  & 5             & 114               \\ \cline{2-6} 
			& Hazelnut   & 391   & 110  & 4             & 136               \\ \cline{2-6} 
			& Meta Nut   & 220   & 115  & 4             & 132               \\ \cline{2-6} 
			& Pill       & 267   & 167  & 7             & 245               \\ \cline{2-6} 
			& Screw      & 320   & 160  & 5             & 135               \\ \cline{2-6} 
			& Toothbrush & 60    & 42   & 1             & 66                \\ \cline{2-6} 
			& Transistor & 213   & 100  & 4             & 44                \\ \cline{2-6} 
			& Zipper     & 240   & 151  & 7             & 177               \\ \hline
			& Total      & 3629  & 1725 & 73            & 1888              \\ \hline
	\end{tabular}}
	\label{table-AD}
\end{table}
\section{Architecture of our CAE}
The architecture of our deep convolutional autoencoder for compressing and reproducing the multi-scale regional representation $f_{\{1:L\}}(\textbf{x})$ is designed as in TABLE~\ref{tab-cae}. It consists of 6 convolutional layers and only contains $1\times{1}$ convolutions and ReLU activations. For the latent feature dimension $c_{d}$, we randomly sample a subset of regional features from the regional feature map and estimate the latent dimension with Principal Component Analysis (PCA) such that 90\% variance is just retained.
\begin{table}[h]
	\centering
	\caption{The architecture of our deep convolutional autoencoder.}
	\resizebox{.8\columnwidth}!{
		\begin{tabular}{l}
			\hline
			\hline
			\textbf{Input}: $f_{\{1:L\}}(\textbf{x})$  $(h_{o}\times{w_{o}}\times{c_o})$ \\ \hline \hline
			[layer 1]: Conv. (1, 1, $(c_{o}+c_{d})//2$), stride=1; ReLU;             \\ \hline
			[layer 2]: Conv. (1, 1, $2\times{c_{d}}$), stride=1; ReLU;             \\ \hline
			[layer 3]: Conv. (1, 1, $c_{d}$), stride=1;              \\ \hline
			[layer 4]: Conv. (1, 1, $2\times{c_{d}}$), stride=1; ReLU;             \\ \hline
			[layer 5]: Conv. (1, 1, $(c_{o}+c_{d})//2$), stride=1; ReLU;         \\ \hline
			[layer 6]: Conv. (1, 1, $c_o$), stride=1;           \\ \hline
	\end{tabular} }
	\label{tab-cae}
\end{table}


\section*{Acknowledgment}
This work is supported by grants from: National Natural Science Foundation of China (No.71932008, 91546201, and 71331005).

\ifCLASSOPTIONcaptionsoff
  \newpage
\fi




\bibliographystyle{IEEEtran}
\bibliography{anomaly_seg}


%
%

%

\begin{IEEEbiography}{Michael Shell}
Biography text here.
\end{IEEEbiography}

\begin{IEEEbiographynophoto}{John Doe}
Biography text here.
\end{IEEEbiographynophoto}


\begin{IEEEbiographynophoto}{Jane Doe}
Biography text here.
\end{IEEEbiographynophoto}




\end{document}